\documentclass{article}

\PassOptionsToPackage{round,authoryear}{natbib}
\usepackage[preprint]{neurips_2021}

\usepackage[T1]{fontenc}
\usepackage[utf8]{inputenc}
\usepackage{booktabs}
\usepackage{array}
\usepackage{microtype}
\usepackage{enumitem}
\usepackage[hidelinks]{hyperref}

\newcommand{\code}[1]{\texttt{#1}}
\newcommand{\na}{\code{Not answerable}}

\title{MMLongBench-Doc-V2: A Corrected-Annotation, Semantics-Aware Revision of
MMLongBench-Doc}
\author{Mingtian Zhang\thanks{Correspondence: \texttt{mingtian@pageindex.ai}}\\
PageIndex}

\begin{document}
\maketitle

\begin{abstract}
MMLongBench-Doc \citep{ma2024mmlongbench} is a long-document QA benchmark of 1{,}082 questions over
135 PDFs. Two properties
of it push measured scores away from the quantity they are meant to capture: the reference metric
compares extracted answer \emph{strings}, so \code{1,358,000} loses to \code{1358000}; and a
non-trivial share of ground-truth annotations are wrong, ambiguous, or incomplete --- concentrated,
because of how they were found, in exactly the questions capable systems answer correctly.
MMLongBench-Doc-V2 corrects 106 annotations, each published with the page and arithmetic that settle
it, and replaces the string metric with a pinned LLM judge asked whether a response \emph{means} the
reference. Ten questions whose document ships under the wrong filename are removed rather than counted
wrong, along with one duplicated question, leaving 1{,}071 questions over 134 documents.
The most reusable contribution is a decision procedure for when an ``empty set'' key may be widened
and when widening would destroy a deliberate negative sample; applied to all 208 \na{} rows,
it widened 14. V2 scores are not comparable with published V1 numbers. The corrected corpus, the
per-entry correction record and the evaluation harness are available at
\url{https://github.com/VectifyAI/MMLongBench-Doc-V2}.
\end{abstract}

\section{Introduction}

MMLongBench-Doc \citep{ma2024mmlongbench} asks questions over 135 PDFs averaging 47.5 pages and
21{,}214 tokens, drawn from seven domains --- research reports, academic papers, guidebooks,
tutorials and workshop decks, financial filings, product brochures, and administrative files. Each
question carries a short, deterministic reference answer, and the evidence needed to produce it is
spread both across pages and across modalities: body text, layout, tables, charts, and figures.
Measured on the distributed annotations, 33.3\% of questions are cross-page, and 20.6\% are
deliberately unanswerable --- included to suppress shortcuts and to detect hallucination directly
rather than inferring it. In the original evaluation of 14 large vision-language models, the strongest
(GPT-4o) reached an overall F1 of 44.9\%, so the headroom is genuine and not an artefact of a saturated
task.

That unanswerable design is why we adopted this benchmark over the alternatives: it is one of the few
document-QA suites where confidently answering a question the document does not support costs a system
points. It is also why the benchmark is worth repairing rather than replacing. Two properties get in
the way of using it as a measurement instrument.

\textbf{The metric measures string identity, not correctness.} V1 extracts a short answer from the
response with GPT-4o, then compares that string to the key, so every formatting decision becomes a
scoring event: \code{Operating Activities} loses to \code{Operations activities}, \code{1358000} to
\code{1,358,000}, \code{-60.3\%} to \code{60.3\% decrease}. The penalty is not uniform --- it is
heaviest where answers are lists or free text --- so it distorts the \emph{relative} ranking of
systems, not merely the absolute level. Six questions are in fact unscoreable no matter what a system
answers, because their own reference answer does not parse under their declared \code{answer\_format}
(\code{int("21\%")} raises; an empty list under \code{List} raises \code{IndexError}).

\textbf{Some annotations are wrong, and not at random.} We found them by auditing questions that a
document-QA system answered confidently and therefore scored zero --- and a meaningful fraction of
those zeros were the system being right, or at least as defensible as the key. Errors of this kind are
invisible to a weak system, which gets the question wrong anyway, and maximally costly to a strong
one. They impose a ceiling on the benchmark that is indistinguishable, from outside, from a ceiling on
the models.

MMLongBench-Doc-V2 addresses both. Section~\ref{sec:metric} describes the replacement metric,
Section~\ref{sec:corrections} the 106 corrections and the annotation failure modes they fall into, and
Section~\ref{sec:emptyset} a decision procedure for the one class of key that cannot be corrected by
inspection alone --- an ``empty set'' answer, where widening the key and preserving a deliberate
negative sample pull in opposite directions. Section~\ref{sec:stats} reports the resulting corpus and
Section~\ref{sec:limitations} what the audit does not establish.

\section{The metric}
\label{sec:metric}

Each row is judged once. The judge receives the question, the reference answer, the expected format,
and the system's \textbf{full response text} --- not a pre-extracted short answer --- and returns a
binary equivalence verdict plus an \code{abstained} flag. Accuracy, recall, precision and F1 keep
V1's definitions; the one substitution is that abstention is read off the response itself rather than
by testing \code{pred == "Not answerable"} on an extractor's output. Precision therefore counts a
confident wrong answer to an unanswerable question against the system, which is what stops an
always-guess policy from scoring well.

Three properties make an LLM judge a metric rather than a new source of noise:

\begin{itemize}
\item \textbf{It never reads the document.} It is given the reference, told to treat it as correct,
  and asked only whether the response says the same thing. This is deliberate. A document-reading
  judge fails on image-only PDFs, truncated text layers and mojibake, and then reports fabrication
  that did not happen; a judge with no document access has nothing it can fail to see, and cannot
  invent a new correct answer.
\item \textbf{The rubric is explicit.} Wording, case, units, separators, list order and surrounding
  prose are free. A different value, a list with missing or extra members, a decline, or a shotgun
  enumeration that merely \emph{contains} the reference are not. For \na{} references only a clean
  decline counts --- one that then volunteers a nearest figure has supplied an answer.
\item \textbf{It is pinned.} A fixed judge model at fixed reasoning effort with a strict JSON schema,
  reported alongside any score and overridable so judge sensitivity can itself be measured.
\end{itemize}

Because the metric changed, \textbf{V2 numbers are not comparable with published V1 numbers}: they are
higher for the same system, and the gap is largest where answers are lists or free text.

\section{Corrections}
\label{sec:corrections}

106 of V1's 1{,}082 annotations are changed; the remaining 976 are byte-identical to upstream. Each
change ships with the original annotation, the correction, and a \code{note} quoting the page or
showing the arithmetic, so a reader can disagree with the reasoning rather than only the verdict.
Upstream's original file ships alongside for independent diffing. Fields touched: \code{answer} 64,
\code{question} 38, \code{evidence\_pages} 27, \code{answer\_format} 19.

\begin{table}[h]
\centering\footnotesize
\caption{The 106 corrections by \code{dispute\_type}.}
\begin{tabular}{@{}lr>{\raggedright\arraybackslash}p{9.0cm}@{}}
\toprule
\code{dispute\_type} & $n$ & Meaning \\
\midrule
\code{wrong\_answer}            & 26 & The question is sound, but the recorded answer is not what the document says. \\
\code{defective\_question}      & 19 & The question is broken --- wrong page, wrong year, a typo, a false premise, or singular phrasing over a multi-part answer. V2 rewords it and keeps the answer. \\
\code{ambiguous\_question}      & 19 & Two readings are both defensible and the answer fits only one. V2 states the intended reading in the question. \\
\code{incomplete\_answer}       & 16 & The key omits members, or omits an equally correct form of the answer. \\
\code{should\_be\_answerable}   & 14 & Labelled \na, but the document contains a determinate answer. \\
\code{corpus\_defect}           & 10 & No answer is recoverable from the distributed corpus. Removed from the data file rather than counted wrong. \\
\code{should\_be\_unanswerable} &  2 & An answer is recorded, but the document does not support one. \\
\bottomrule
\end{tabular}
\end{table}

\textbf{38 corrections change the question, not the answer.} Where a document genuinely supports two
readings, pinning the intended one in the question is more honest than declaring one reading wrong. A
question asking for ``total debt'' in a filing that reports short-term borrowings, current maturities,
long-term debt and lease liabilities separately does not have one answer; it has a family of answers
indexed by a convention the annotator held silently. But \emph{stating} a convention can introduce a
new one: adding ``all debt plus all lease liabilities'' fixed the lease ambiguity and immediately
created another, because \$41M of short-term borrowings disclosed only in a note is literally ``all
debt''. The final wording names the balance-sheet lines. Prefer the most specific phrasing available;
a universal quantifier moves the boundary rather than removing it.

\textbf{\code{corpus\_defect} is not a hard question.} Ten questions target a document that ships
under the wrong filename --- the distributed PDF is a different document from the one the questions
were written against. No system can answer them and no key is recoverable, so counting them wrong
measures the corpus, not the system. They are dropped from the data file; all ten belong to one
document, so V2 covers 134 of upstream's 135 PDFs. \code{corrections.json} keeps them as the record
of why they were removed.

\textbf{Recurring failure modes.} Most corrections fall into five patterns, each worth checking for in
any similar corpus: wrapped table cells counted as multiple entries (a troubleshooting table keyed 17
where it lists 15); a survey percentage multiplied by the wrong base (a ``\% of internet users''
figure applied to every respondent, giving counts larger than the qualifying population); off-by-one
between chart gridlines and printed labels; values transcribed from an adjacent row --- the hardest to
spot, since the result is plausible and internally consistent; and category confusions, including one
question asking for ``interest coverage ratio for AMCOR FY2020'' inside a Best Buy 10-K where the
recorded value is Best Buy's own FY2023 figure. A representative correction: the percentage change of
Amazon's return allowance from 2016 to 2017 was recorded as \code{60.3\%}, but the allowance
\emph{fell} from 156 to 62, so the change is \code{-60.3\%} --- the magnitude was right and the sign
was dropped.

\section{When an empty-set key may be widened}
\label{sec:emptyset}

MMLongBench-Doc keys the same question shape two ways. Of the 287 upstream questions beginning ``how
many'', \textbf{14 are keyed \code{0}} for a thing simply absent --- cats, tigers, airplanes, blue
arrows, GPT-4o, words starting with `X' --- while \textbf{56 are keyed \na}, including questions of
essentially identical shape. One brochure contradicts itself inside a page: ``how many dogs and cats
are there in page 17'' is keyed \code{['0','0']} and ``how many people with scarf are there in Page
5'' is keyed \code{2}, but the sunglasses twin on that same page is keyed \na.

This cuts both ways, which makes it a real problem rather than a stylistic one: a system answering
``there are none'' is wrong on the \na{} rows, and one that declines is wrong on the \code{0} rows. No
policy scores well, and the choice between them is unrelated to document understanding. Fourteen keys
were therefore widened to \code{Not answerable or 0 or none}, with the metric matching the \na{}
prefix so widened rows stay in the unanswerable slice. All 208 \na{} rows were reviewed
against the following test, which turns on \textbf{what exactly is missing}.

\textbf{Widen} when the container the question names \emph{exists} and the thing asked about is
\emph{verifiably absent from it}. Then ``none'' --- or, for a \code{how many}, \code{0} --- is a
reading of the document, not a guess. Verified examples: a street photograph contains three cars and a
box truck and no bicycle; two named pages contain zero raster images and zero large vector blocks, so
there is no diagram on that page under either page numbering; a timer chart draws only green and blue
bars, so no red bar starts anywhere.

\textbf{Do not widen} in the four cases below, where ``none'' is an over-claim rather than a reading.

\begin{table}[h]
\centering\footnotesize
\caption{The four cases where an empty-looking key must be left alone.}
\begin{tabular}{@{}>{\raggedright\arraybackslash}p{4.5cm}>{\raggedright\arraybackslash}p{9.0cm}@{}}
\toprule
Case & Example \\
\midrule
The container itself is missing
  & Page ranges outside the document (\code{Pages 200--205}, \code{Pages 400--640}); a ``Figure 10''
    with no caption anywhere in the paper. \\
The entity never appears
  & Beijing in a Nepal media survey; a microwave in a dishwasher manual; a ``level-6'' in a
    five-level system; survey years the report does not cover. \\
The question asks for a complement against an unbounded universe
  & ``signal icons that can \emph{not} be found in the Status Bar'' --- the set of absent icons is not
    enumerable, so ``none'' is not a claim the document licenses. \\
The set is short, not empty
  & A deck is asked for ``another \emph{two} companies'' when exactly one further contact carries a
    phone number. \\
\bottomrule
\end{tabular}
\end{table}

The last row is why this is a test and not a heuristic: that key survived two passes of review looking
like an empty set, and what caught it was mechanical --- an assertion that the candidate set had zero
members found one. \textbf{Gate every empty-set widening on an executable emptiness check}, not on the
reviewer's reading. The failure mode is silent: a wrongly widened key accepts a wrong answer forever
and nothing downstream flags it.

\section{The resulting corpus}
\label{sec:stats}

V2 contains \textbf{1{,}071 questions over 134 documents}; 208 (19.4\%) are unanswerable, 14 of them
with a widened key. V1's 1{,}082 becomes 1{,}071 because ten rows are dropped as described above, and
because one
question appears twice under the same \code{doc\_id} with contradictory answers --- once \code{46\%},
once \na. The document answers it on page 97, so both resolve to \code{46\%}, leaving a duplicate
\code{(doc\_id, question)} key, which silently drops a row from any keyed join; the second copy is
removed. The file keeps upstream's name and schema, so V1 tooling reads it unchanged.

\begin{table}[h]
\centering\footnotesize
\caption{Corpus composition. Task-type and \code{requires\_visual} labels are model-produced and not
human-verified; everything else is upstream metadata carried through unchanged.}
\begin{tabular}{@{}lr@{\hskip 1.6em}lr@{}}
\toprule
Document type & $n$ & Task type & $n$ \\
\midrule
Research report / Introduction & 281 & \code{lookup}    & 487 \\
Academic paper                 & 199 & \code{count}     & 234 \\
Guidebook                      & 155 & \code{enumerate} & 122 \\
Tutorial / Workshop            & 138 & \code{derive}    & 122 \\
Financial report               & 117 & \code{compare}   &  89 \\
Brochure                       & 100 & \code{verify}    &  17 \\
Administration / Industry file &  81 & \multicolumn{2}{@{}l}{} \\
\midrule
\multicolumn{2}{@{}l}{\emph{Evidence source}} & \multicolumn{2}{@{}l}{\emph{Evidence location}} \\
Figure                     & 302 & Single page         & 496 \\
Pure-text (plain text)     & 299 & Cross-page          & 363 \\
Table                      & 216 & None (unanswerable) & 212 \\
Chart                      & 173 & & \\
Generalized-text (layout)  & 119 & & \\
\bottomrule
\end{tabular}
\end{table}

A per-question \code{requires\_visual} flag marks the 314 questions needing a property plain-text
extraction cannot carry --- colour, shape, layout, an icon's presence, the content of a photograph.
\textbf{This is not the same as \code{evidence\_sources} containing \code{Chart} or \code{Figure}}, and
the gap runs both ways: of the questions marked \code{Chart}/\code{Figure}, 231 are answerable
straight from the text layer (the figure is an image, but the number wanted is printed as a chart
label that extraction picks up), while 75 questions \emph{not} so marked do need the rendered page ---
typically meta-questions like ``how many bar charts are in the report''. Slicing a ``visual'' subset on
\code{evidence\_sources} alone therefore mixes in many pure-text questions and misses genuinely visual
ones.

\section{Limitations}
\label{sec:limitations}

\textbf{The corrections come from a biased slice} --- questions one system answered confidently and
scored zero. That is where annotation errors concentrate, which makes it efficient, and it means the
corpus-wide error rate \textbf{remains unmeasured}; a uniform random audit would establish it and we
have not run one. \textbf{Coverage is uneven by task type}: \code{lookup}, \code{derive},
\code{compare} and \code{verify} are audited end to end, \code{count} and \code{enumerate} are not, so
their error rates are the least trustworthy --- and they are also the hardest slices, resisting
verification from the text layer alone. Roughly 95 rows in those two categories are ones where a
capable system disagrees with the key and we could not prove the key wrong; they are documented but
unchanged. \textbf{Thirteen entries record the change but not the reasoning}; they come from the first
review pass, were checked against the cited pages at the time, and their \code{note} says so rather
than offering a post-hoc justification. \textbf{The task-type and \code{requires\_visual} labels are
model-produced} --- one pass, one model, low effort, never human-verified; treat them as a convenience
for slicing, not ground truth. \textbf{The judge is a model}: pinned and explicitly ruled, but not
infallible, and judge choice is a free parameter of the metric, so report the judge model and provider
with any score.

\section{Availability}

The corrected data file, the per-entry correction record, upstream's original file for independent
diffing, and the evaluation harness are released under Apache~2.0, inherited from upstream, with this
revision tagged \code{v2.0}. The source PDFs are \textbf{not} redistributed and should be obtained
upstream; nothing in the harness opens them, as scoring only ever sees the question, the reference
answer, and the response. Corrections are welcome --- the useful form is an issue naming the entry
and the page that settles it.


\small
\begin{thebibliography}{1}
\bibitem[Ma et al.(2024)]{ma2024mmlongbench}
Yubo Ma, Yuhang Zang, Liangyu Chen, et al.
\newblock MMLongBench-Doc: Benchmarking long-context document understanding with visualizations.
\newblock \emph{arXiv preprint arXiv:2407.01523}, 2024.
\newblock URL \url{https://arxiv.org/abs/2407.01523}.
\end{thebibliography}
\end{document}